\documentclass{article}
\usepackage[preprint]{spconf}
\usepackage{amsmath, graphicx}
\usepackage{booktabs}
\usepackage{hyperref}

\usepackage[ruled]{algorithm2e} 

\title{Incorporating Eye-Tracking signals into Multimodal Deep Visual Models for Predicting User Aesthetic Experience in Residential Interiors}
\twoauthors
  {Chen-Ying Chien}
	{Institute of Information Systems and Applications\\
	National Tsing Hua University\\
	Hsinchu, Taiwan}
  {Po-Chih Kuo}
	{Department of Computer Science\\
	National Tsing Hua University\\
	Hsinchu, Taiwan}
\begin{document}

%
\maketitle
\begin{abstract}
Understanding how people perceive and evaluate interior spaces is essential for designing environments that promote well-being. However, predicting aesthetic experiences remains difficult due to the subjective nature of perception and the complexity of visual responses. This study introduces a dual-branch CNN-LSTM framework that fuses visual features with eye-tracking signals to predict aesthetic evaluations of residential interiors. We collected a dataset of 224 interior design videos paired with synchronized gaze data from 28 participants who rated 15 aesthetic dimensions. The proposed model attains 72.2\% accuracy on objective dimensions (e.g., light) and 66.8\% on subjective dimensions (e.g., relaxation), outperforming state-of-the-art video baselines and showing clear gains on subjective evaluation tasks. Notably, models trained with eye-tracking retain comparable performance when deployed with visual input alone. Ablation experiments further reveal that pupil responses contribute most to objective assessments, while the combination of gaze and visual cues enhances subjective evaluations. These findings highlight the value of incorporating eye-tracking as privileged information during training, enabling more practical tools for aesthetic assessment in interior design.
\end{abstract}
\begin{keywords}
Human Building Interaction, AI-aided Design, Spatial Perception, Eye-tracking, Multimodal
\end{keywords}
\section{Introduction}
\label{sec:intro}

Human perception of architectural environments is profoundly influenced by visual processing, which in turn affects occupants’ comfort and emotional responses. Given that individuals spend approximately 90\% of their time indoors \cite{klepeis2001national}, it is critical to understand how interior spaces impact their experiences. Visual attention has been shown to correlate with spatial preferences and emotional states \cite{tuszynska2020effects}, rendering gaze behavior a valuable indicator in this context.

Traditionally, the design of interior spaces has relied on designers’ professional expertise, often without empirical evaluation of occupants’ experiences. When user feedback is incorporated, it is commonly obtained through questionnaires and interviews, which suffer from recall bias and linguistic limits. While neuroscientific methods like electroencephalography (EEG) and functional magnetic resonance imaging (fMRI) yield detailed neural insights \cite{coburn_psychological_2020}, their high costs and laboratory constraints limit their practical applicability. In contrast, eye-tracking technology offers a more accessible and objective approach to capturing real-time visual attention, thereby overcoming the drawbacks associated with self-report and neuroimaging techniques \cite{de2020subjective}. Nevertheless, the potential of eye-tracking for predicting aesthetic evaluations of interior spaces remains underexplored, particularly in dynamic video settings with multidimensional user responses.

In this study, we present a dual-branch CNN–LSTM framework that integrates eye-tracking with video to predict users' aesthetic evaluations of interior spaces. Our contributions are:  
(1) a multimodal eye-tracking dataset;  
(2) a dual-branch CNN–LSTM model that fuses visual and gaze features for spatio-temporal modeling;  
(3) a practical system that requires gaze only during training, enabling deployment from video alone.

\section{Related Works}
\label{sec:related_works}

Eye-tracking records eye movements to reveal visual attention and cognitive processes.  
Early research focused on predicting where and what people look at, advancing saliency and gaze-prediction models \cite{xu2018gaze, chen2023deep} but addressing only spatial attention.

More recent studies link gaze behavior with subjective experience.  
For instance, Christoforou et al.\ \cite{christoforou2015eyes} predicted advertising video preferences based on gaze metrics, while additional studies have established associations between fixation duration, scan-path complexity, and the distribution of attention and aesthetic judgments \cite{kwon2021meaning}.

Within architectural research, eye-tracking has been used to associate spatial design elements with emotional states.
Tuszyńska-Bogucka et al. \cite{tuszynska2020effects} reported that pupil dilation distinguishes positive from negative reactions to interior environments, while Kim and Lee \cite{kim2021assessing} further identified correlations between gaze behavior and emotional arousal in virtual retail settings.

Despite these advances, most studies have focused on static images, single-room settings, or virtual reality environments requiring specialized equipment.  
Video, a medium increasingly used to present interior design, remains comparatively underexplored.  
This study addresses that gap by analyzing eye-tracking data collected during interior-design video viewing to predict multidimensional aesthetic evaluations.

\section{Dataset}
\label{sec:dataset}

\subsection{Participants}
A total of thirty adults, aged between 18 and 32 years, were initially recruited for the study. Following quality control procedures \cite{holmqvist2011eye}, two participants were excluded, resulting in a final sample of twenty-eight valid participants with balanced gender representation. 
The study received approval from the institutional review board.

\subsection{Stimuli and Apparatus}
A total of sixteen first-person walkthrough videos (80 seconds in duration, 1920×1080 resolution, 30 frames per second) were produced, each illustrating four distinct Taiwanese townhouse layouts rendered in four design styles: Modern, Nordic, Wabi-Sabi, and MUJI. These videos were standardized with consistent lighting conditions set at 2 p.m. and a fixed field of view of 100° \cite{lin2025shaping}. Eye movement data were captured using a Tobii Pro Spark eye tracker operating at 60 Hz, with stimuli displayed on a 27-inch monitor. The presentation of stimuli and the collection of responses were conducted utilizing PsychoPy software.

\subsection{Procedure}
The participants viewed eight videos selected at random, during which their eye movements were continuously recorded. Upon completion of each video, participants were asked to provide a concise verbal description, answered two lighting-perception questions \cite{chinazzo2021temperature}, and rated sixteen aesthetic dimensions on 7-point Likert scales \cite{coburn_psychological_2020} (full questionnaire in \cite{lin2025shaping}). 

\section{Methodology}
\label{sec:method}

\subsection{Overview}
Our proposed system integrates synchronized video and eye-tracking data within a unified predictive framework. Initially, the model is trained using both modalities to learn the relationship between visual behavior and aesthetic evaluations, with validation performed on the same inputs. Subsequently, the model is deployed relying solely on video data, treating gaze information as privileged knowledge that facilitates training but is not required during inference. 



\subsection{Data Preprocessing}
To reduce subject-specific rating bias, participant scores are normalized on a per-subject basis.  
Eye-tracking signals are cleaned and converted into two representations:  
(1) pupil-response images, generated from task-evoked pupillary responses using Gramian Angular Field and Markov Transition Field transforms \cite{sharma2021multi}; and  
(2) visual-attention maps, produced by aggregating gaze coordinates in 1-second windows and applying Gaussian blur ($\sigma \approx 1^\circ$ visual angle) \cite{holmqvist2011eye}.
All representations are temporally aligned with the video frames.

\subsection{Model Architecture}

\begin{figure*}[t]
\centering
\includegraphics[width=0.85\linewidth]{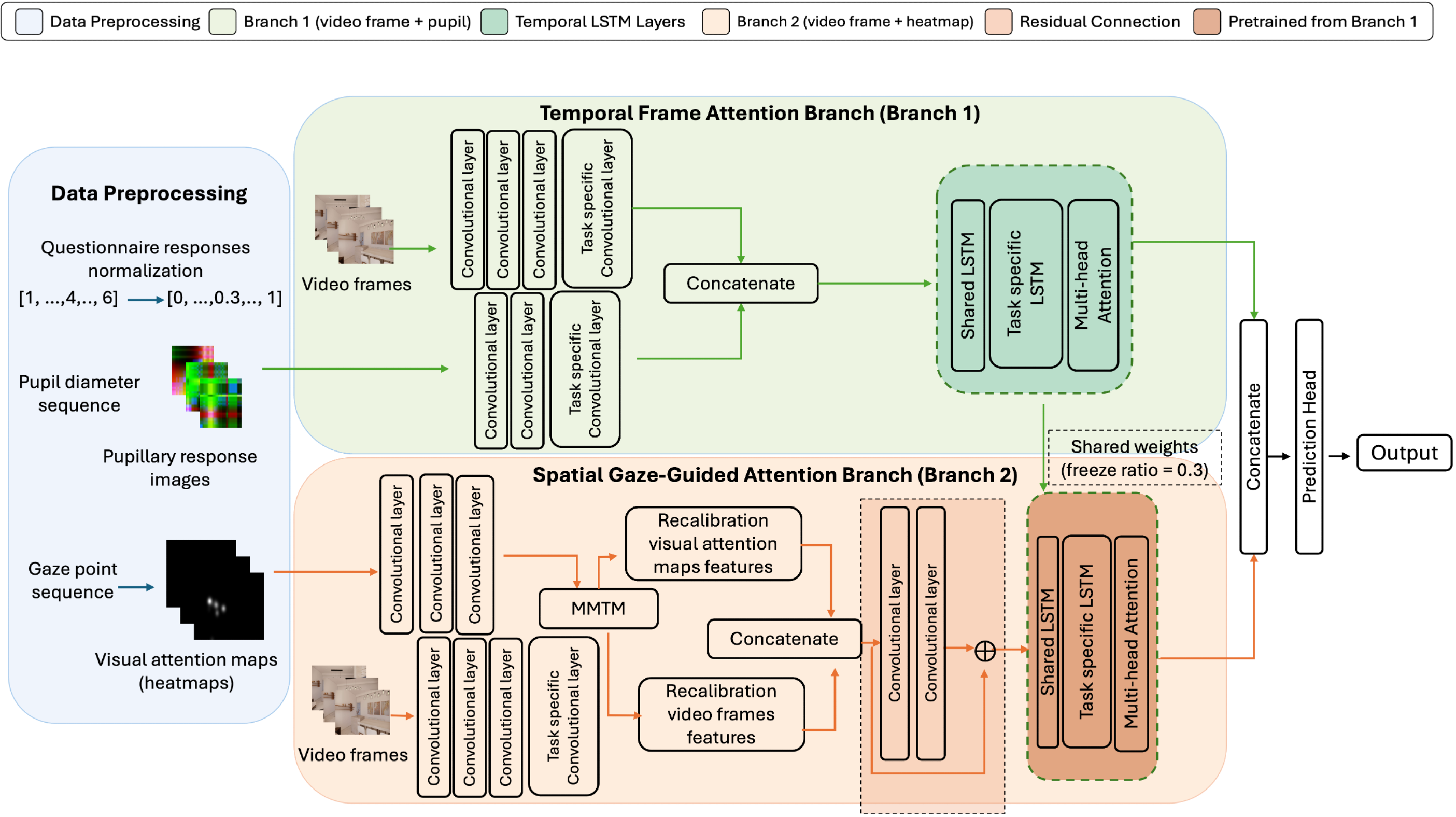}
\caption{Architecture of the proposed dual-branch CNN–LSTM model.}
\label{fig:model_overview_with_label}
\end{figure*}

The proposed architecture comprises two complementary branches, as illustrated in Fig. ~\ref{fig:model_overview_with_label}. 

\textbf{Temporal Frame Attention Branch.}
This branch is designed to capture temporal viewing dynamics by jointly processing video frames alongside their associated pupil-response images. The video pathway initially passes through a shared three-layer convolutional backbone, consisting of convolution, batch normalization, ReLU activation, and max-pooling layers, followed by a task-specific convolutional layer tailored to each aesthetic question. Concurrently, the pupil-response pathway employs a two-layer shared convolutional encoder succeeded by a task-specific convolutional layer. The resulting feature representations from both streams are concatenated and subsequently input into shared LSTM layers to model temporal dependencies across the entire 80-second sequence.

\textbf{Spatial Gaze-Guided Attention Branch.}
This branch focuses on modeling spatial attention by processing video frames and gaze-derived visual-attention maps through parallel CNN streams. The video stream replicates the three-layer shared plus task-specific structure utilized in the Temporal Frame Attention Branch, whereas the attention-map stream comprises a three-layer shared backbone exclusively. Cross-modal interactions are facilitated by a Multimodal Transfer Module (MMTM) \cite{joze2020mmtm}, which performs global average pooling, feature concatenation, and channel-wise recalibration. The recalibrated features are integrated via residual connections and then passed through shared and task-specific LSTM layers to capture temporal context while maintaining pre-fusion information integrity.

Finally, the outputs from both branches are concatenated and processed through two fully connected layers with sigmoid activation functions to generate multi-label binary predictions corresponding to each aesthetic dimension.

\subsection{Training Strategy}
To give the Spatial branch a stable temporal prior, we train the network in three stages.  
(1) Temporal pretraining: the Temporal Frame Attention Branch is trained first on video and pupil data until convergence.  
(2) Weight transfer: the hidden-to-hidden LSTM weights learned in Stage 1 initialize the Spatial Gaze-Guided Attention Branch, while its input-to-hidden weights are freshly initialized to match the different feature size.  
(3) Joint fine-tuning: both branches are then trained together, with roughly 30\% of the transferred weights frozen to preserve the learned dynamics. 
We optimize the full model using Adam and binary cross-entropy loss, reporting accuracy for all 15 aesthetic dimensions.

\section{Experimental Results}
\label{sec:results}

\subsection{Experimental Setup}
Due to the limited variability observed across the video stimuli concerning three aesthetic dimensions (ugly–beautiful, impersonal–personal, and aged–modern), these dimensions were excluded from subsequent modeling. Consequently, each video was associated with 15 prediction tasks. The resulting dataset comprises 224 interior-design video sequences accompanied by synchronized eye-tracking data. The dataset was partitioned at the participant level into training (70\%), validation (15\%), and test (15\%) sets.

\subsection{Baseline Comparison}
Since no existing work employs similar methods for aesthetic evaluation, we evaluate our approach against five state-of-the-art video understanding models: I3D~\cite{carreira2017quo}, X3D~\cite{feichtenhofer2020x3d}, Video Swin Transformer~\cite{liu2022video}, TimeSformer~\cite{bertasius2021space}, and VideoMamba~\cite{li2024videomamba}. The outcomes, summarized in Tables~\ref{tab:objective_results_baseline} and \ref{tab:subjective_results_baseline}, report accuracy metrics on both objective and subjective question sets. Our proposed model achieves an accuracy of 72.2\% on objective tasks and 66.8\% on subjective tasks. 

\begin{table}[t]
\centering
\scriptsize
\setlength{\tabcolsep}{3pt}
\renewcommand{\arraystretch}{0.9}
\begin{tabular}{lccccc}
\toprule
\textbf{Method} & \textbf{Overall} &
\textbf{Light} &
\textbf{Complexity} &
\textbf{Organization} &
\textbf{Naturalness} \\
\midrule
I3D~\cite{carreira2017quo} & 0.572 & 0.571 & 0.486 & \underline{0.743} & 0.629 \\
X3D~\cite{feichtenhofer2020x3d} & 0.629 & 0.571 & 0.314 & \underline{0.743} & 0.629 \\
Video Swin~\cite{liu2022video} & 0.721 & \underline{0.714} & 0.657 & \textbf{0.771} & \underline{0.714} \\
TimeSformer~\cite{bertasius2021space} & \textbf{0.743} & \textbf{0.743} & \underline{0.714} & \textbf{0.771} & \textbf{0.743} \\
VideoMamba~\cite{li2024videomamba} & 0.615 & 0.600 & 0.600 & \underline{0.743} & 0.629 \\
\midrule
\textbf{Ours} & \underline{0.722} & \textbf{0.743} & \textbf{0.743} & \underline{0.743} & 0.657 \\
\bottomrule
\end{tabular}
\caption{Accuracy on objective dimensions. \textbf{Bold} indicates best performance, \underline{underline} second best.}
\label{tab:objective_results_baseline}
\end{table}

\begin{table*}[t]
\centering
\scriptsize
\setlength{\tabcolsep}{3pt}
\renewcommand{\arraystretch}{0.9}
\begin{tabular}{lccccccccccccc}
\toprule
\textbf{Method} & \textbf{Overall} & \textbf{Color Comfort} & \textbf{Interest} & \textbf{Valence} &
\textbf{Stimulation} & \textbf{Vitality} & \textbf{Comfort} & \textbf{Relaxation} &
\textbf{Hominess} & \textbf{Uplift} & \textbf{Approachability} & \textbf{Explorability} \\
\midrule
I3D~\cite{carreira2017quo} & 0.516 & 0.486 & 0.486 & 0.514 & 0.543 & 0.476 & 0.600 & 0.457 & 0.571 & 0.437 & 0.571 & 0.514 \\
X3D~\cite{feichtenhofer2020x3d} & 0.525 & 0.314 & 0.543 & \underline{0.571} & 0.543 & 0.514 & 0.600 & 0.400 & 0.629 & 0.600 & 0.543 & 0.514 \\
Video Swin~\cite{liu2022video} & \underline{0.649} & 0.571 & \textbf{0.686} & \textbf{0.629} & 0.571 & \underline{0.629} & \underline{0.629} & \textbf{0.714} & \underline{0.686} & \underline{0.629} & \textbf{0.714} & \textbf{0.686} \\
TimeSformer~\cite{bertasius2021space} & 0.600 & 0.457 & 0.543 & 0.514 & \underline{0.600} & 0.600 & \underline{0.629} & \underline{0.657} & \underline{0.686} & 0.600 & \textbf{0.714} & 0.600 \\
VideoMamba~\cite{li2024videomamba} & 0.532 & \underline{0.600} & 0.514 & 0.486 & 0.571 & 0.514 & 0.600 & 0.486 & 0.543 & 0.457 & 0.514 & 0.514 \\
\midrule
\textbf{Ours} & \textbf{0.668} & \textbf{0.629} & \underline{0.657} & \textbf{0.629} & \textbf{0.657} & \textbf{0.657} & \textbf{0.743} & \underline{0.657} & \textbf{0.714} & \textbf{0.686} & \underline{0.686} & \underline{0.629} \\
\bottomrule
\end{tabular}
\caption{Accuracy on subjective dimensions. \textbf{Bold} indicates best performance, \underline{underline} second best.}
\label{tab:subjective_results_baseline}
\end{table*}


Furthermore, our method demonstrates superior performance on seven out of eleven subjective questions and consistently excels in affective dimensions such as \emph{comfort}, \emph{hominess}, and \emph{uplift}. These findings substantiate the premise that incorporating eye-tracking data yields particularly informative signals for affective assessment.

\subsection{Ablation Study and Model Deployment}
To assess the role of each eye-tracking modality, we removed them during training (Table~\ref{tab:ablation_overall}).  
Excluding pupil data reduced objective accuracy from 72.2\% to 67.9\%, a larger drop than removing gaze heatmaps, indicating that pupil dynamics convey stronger cognitive cues. 
For deployment without eye-tracking, a model trained with full multimodal input but tested on \emph{video only} reached 72.8\% (objective) and 67.0\% (subjective) accuracy, exceeding a video-only baseline.  
This shows that gaze acts as \emph{privileged information}~\cite{vapnik2015learning}, enriching representation learning even when absent at inference.
\begin{table}[t]
\centering
\scriptsize
\setlength{\tabcolsep}{4pt}
\renewcommand{\arraystretch}{0.9}
\begin{tabular}{lcc}
\toprule
\textbf{Method} & \textbf{Objective Dim. Accuracy} & \textbf{Subjective Dim. Accuracy} \\
\midrule
Full model                       & \textbf{0.722} & \textbf{0.668} \\
\midrule
w/o visual attention         & 0.700          & 0.660 \\
w/o pupil                         & 0.679          & 0.667 \\
w/o visual attention \& pupil & 0.679         & 0.649 \\
\bottomrule
\end{tabular}
\caption{Overall accuracy on objective and subjective dimensions under different ablation settings.}
\label{tab:ablation_overall}
\end{table}




\subsection{Attention Visualization}
\begin{figure}[htp]
    \centering
    \includegraphics[width=1\linewidth]{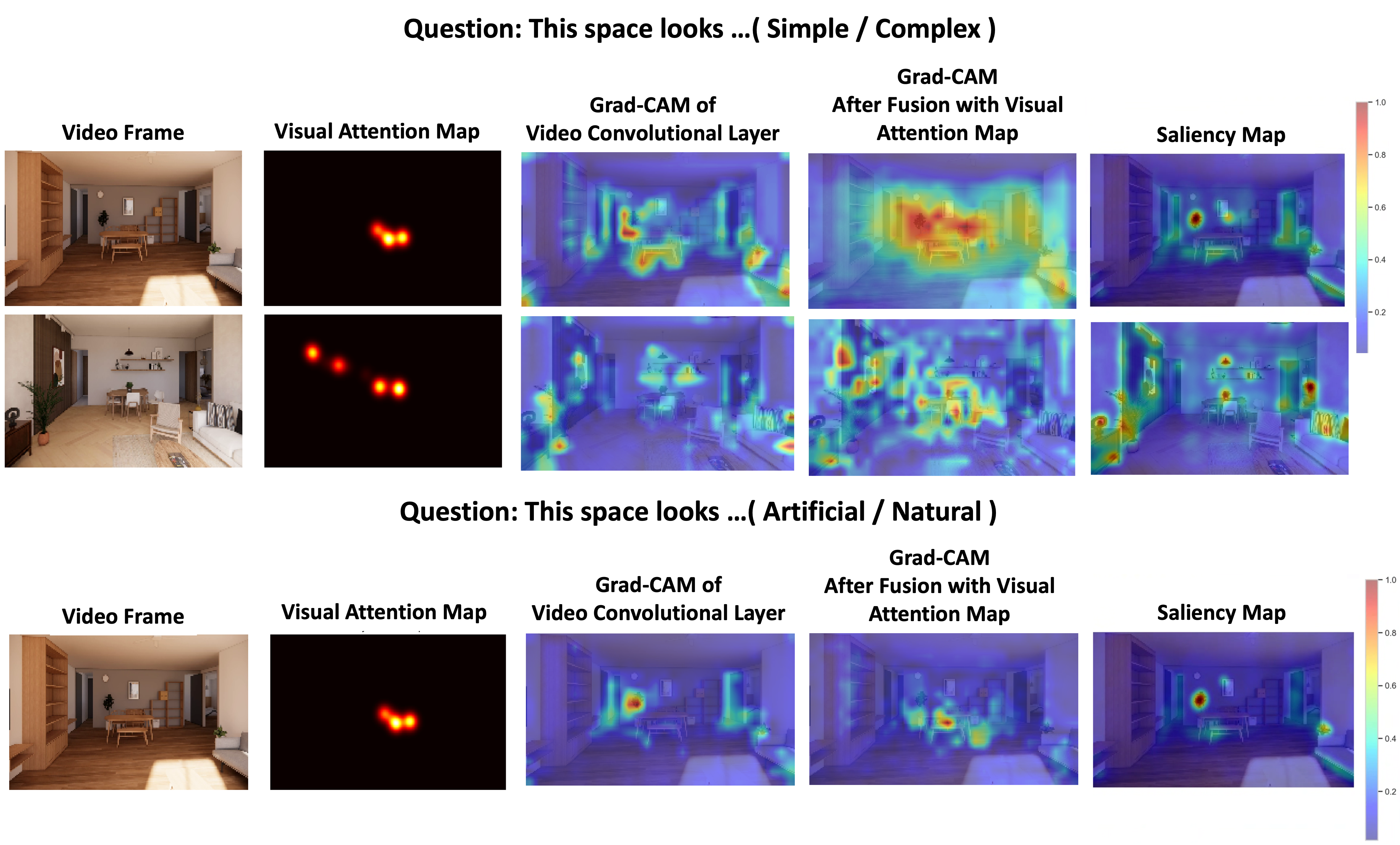}
    \caption{Grad-CAM visualization of task-specific attention. 
    }
    \label{fig:Grad_CAM}
\end{figure}

To examine how guided attention shapes model behavior, we visualize learned activations with Gradient-weighted Class Activation Mapping (Grad-CAM)~\cite{selvaraju2017grad} (Fig.~\ref{fig:Grad_CAM}). For the same interior scene but different evaluation tasks, the model adopts distinct strategies: 
when judging \emph{complexity} it spreads attention across the entire layout to capture global organization, 
whereas for \emph{naturalness} it selectively attends to wooden furniture and plant elements, matching participants’ descriptions of natural materials as cues of naturalness.

\begin{figure}[htp]
    \centering
    \includegraphics[width=1\linewidth]{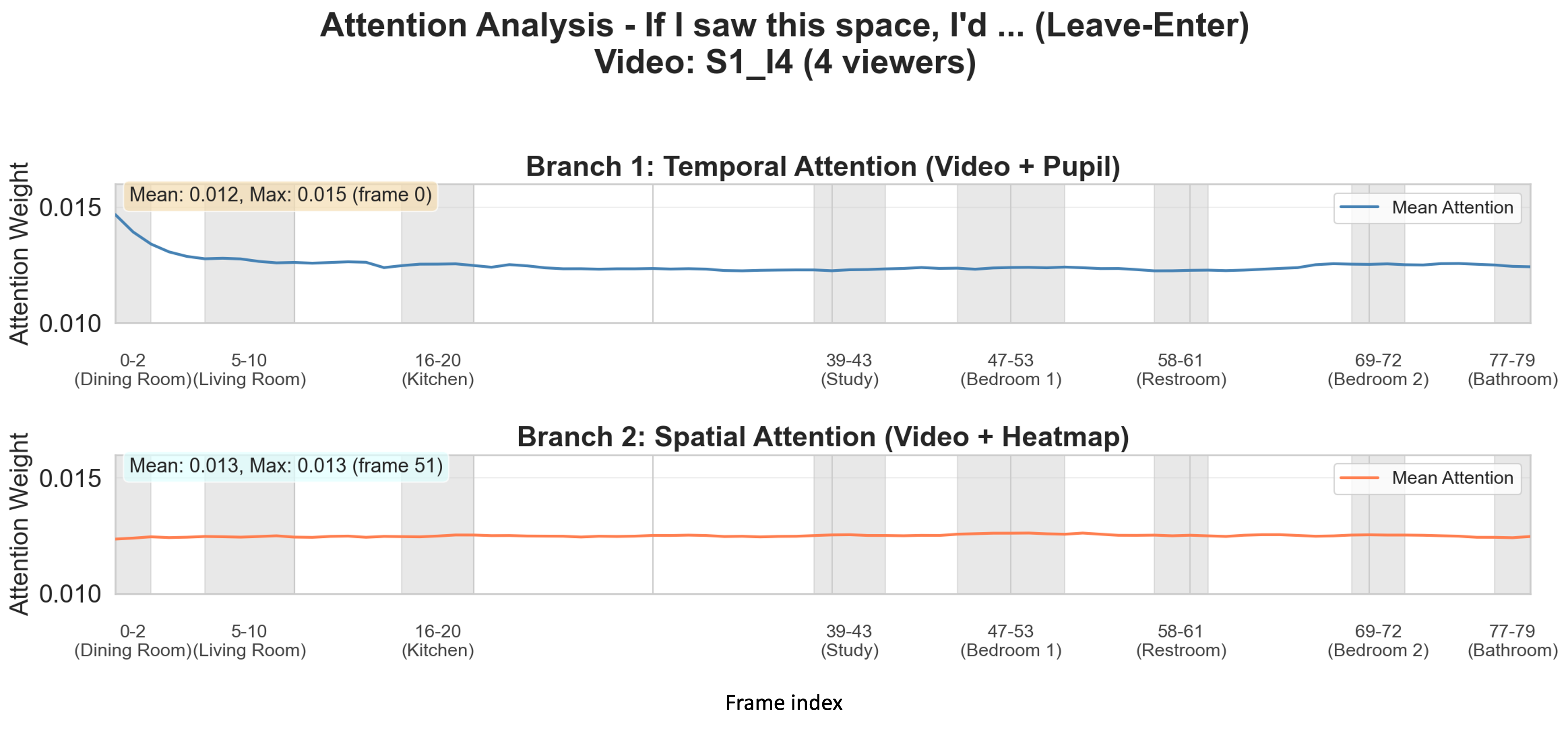}
    \caption{Temporal attention for `Leave-Enter' dimension.}
    \label{fig:temporal_attention_leave_enter}
\end{figure}

\begin{figure}[htp]
    \centering
    \includegraphics[width=1\linewidth]{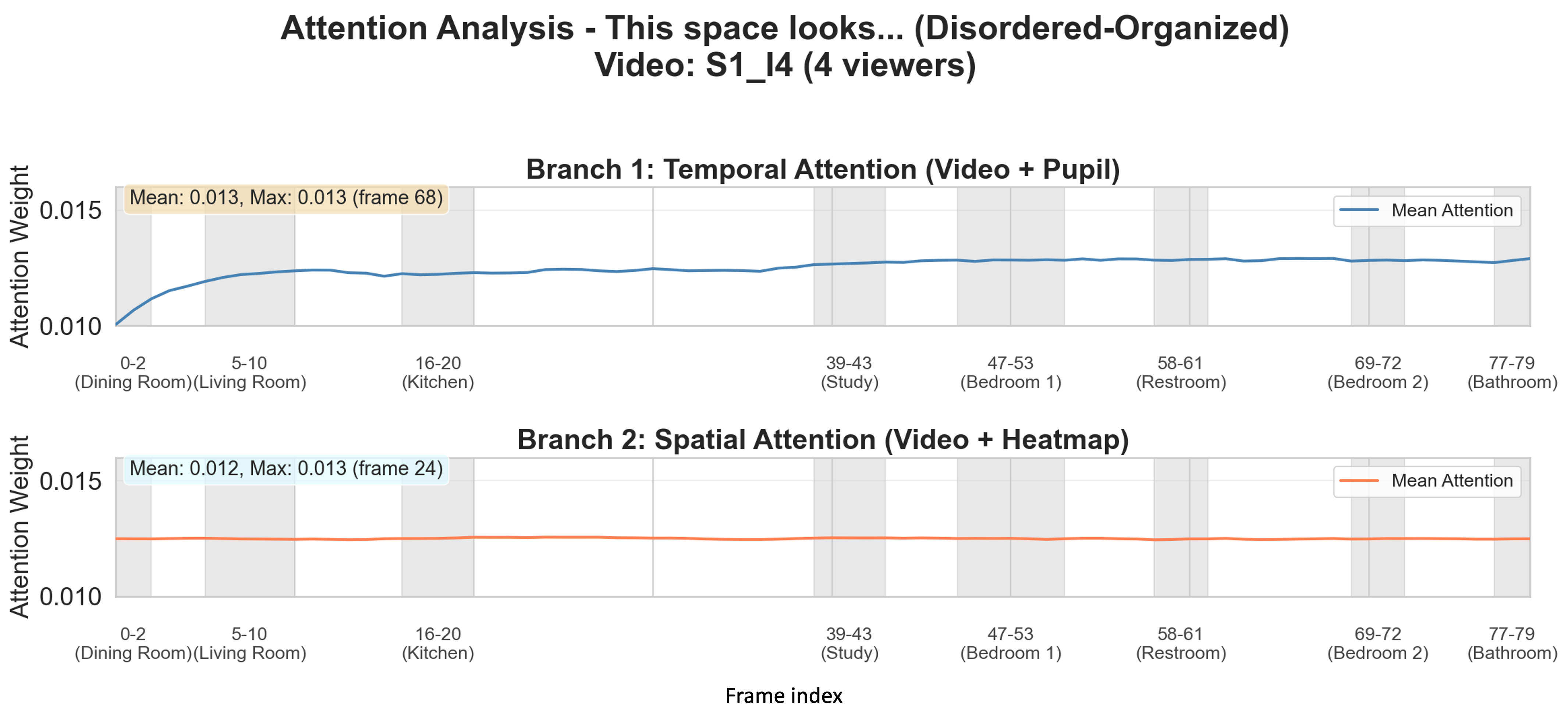}
    \caption{Progressive attention  for organizational assessment.}
    \label{fig:temporal_attention_disordered_organized}
\end{figure}

Temporal attention shows a complementary pattern. 
In behavioral decisions such as “enter/leave,” the Temporal branch exhibits an early peak, reflecting first-impression effects (Fig.~\ref{fig:temporal_attention_leave_enter}), 
whereas for organization-related tasks its attention gradually accumulates throughout the full 80-second sequence (Fig.~\ref{fig:temporal_attention_disordered_organized}). 
These findings confirm that the spatial and temporal branches specialize in stable spatial cues and time-sensitive information, respectively. 

\section{Conclusion}
We proposed a dual-branch CNN–LSTM framework that combines visual content with eye-tracking signals to predict aesthetic evaluations of interior spaces. By treating gaze as privileged information during training, the approach improves model generalization and provides interpretable attention patterns aligned with human perception. This work highlights the potential of integrating eye-tracking to support practical and scalable tools for assessing architectural aesthetics.

{\ninept
\bibliographystyle{IEEEbib}
\bibliography{strings,refs}
}

\end{document}